\documentclass{article}

\usepackage{arxiv}

\usepackage[utf8]{inputenc} % allow utf-8 input
\usepackage[T1]{fontenc}    % use 8-bit T1 fonts
\usepackage{hyperref}       % hyperlinks
\usepackage{url}            % simple URL typesetting
\usepackage{booktabs}       % professional-quality tables
\usepackage{amsfonts}       % blackboard math symbols
\usepackage{nicefrac}       % compact symbols for 1/2, etc.
\usepackage{microtype}      % microtypography
\usepackage{lipsum}

\title{Using Graph Neural Networks for\\Mass Spectrometry Prediction}

\author{
    Hao Zhu
   \\
    Department of Computer Science \\
    Tufts University \\
  Medford, MA, 02153 \\
  \texttt{\href{mailto:hao.zhu@tufts.edu}{\nolinkurl{hao.zhu@tufts.edu}}} \\
   \And
    Liping Liu
   \\
    Department of Computer Science \\
    Tufts University \\
  Medford, MA, 02153 \\
  \texttt{\href{mailto:liping.liu@tufts.edu}{\nolinkurl{liping.liu@tufts.edu}}} \\
   \And
    Soha Hassoun
   \\
    Department of Computer Science \\
    Tufts University \\
  Medford, MA, 02153 \\
  \texttt{\href{mailto:soha.hassoun@tufts.edu}{\nolinkurl{soha.hassoun@tufts.edu}}} \\
  }

\begin{document}
\maketitle

\def\tightlist{}

\begin{abstract}
Detecting and quantifying products of cellular metabolism using Mass Spectrometry (MS) has already shown great promise in many biological and biomedical  applications.  The biggest challenge in metabolomics is \textit{annotation}, where measured spectra are assigned chemical identities. Despite  advances, current methods provide limited annotation for measured spectra.
Here, we explore using graph neural networks (GNNs) to predict the spectra.
 The input to our model is a molecular graph. 
 %; the output is an array of MS  intensities for specified mass-to-charge (m/z) bins.  
 The model is trained and tested on the  NIST 17 LC-MS dataset. We compare our results to NEIMS, a  neural network model that utilizes  molecular fingerprints as inputs. Our results show that  GNN-based models offer higher performance than NEIMS. Importantly, we show that ranking results heavily depend on the candidate set size and on the similarity of the candidates to the target molecule, thus highlighting the need for consistent, well-characterized evaluation protocols for this domain.

%Performance is evaluated using: (1) similarity of the predicted and target %spectrum, and (2) percentage of correctly identified molecules in the test %dataset when considering the top k ranked candidate molecules for each test %spectra.  

\end{abstract}

\hypertarget{introduction}{%
\section{Introduction}\label{introduction}}

Metabolomics is a powerful approach  that can characterize small molecules
produced in cells, tissues and other biological systems. Metabolites are
direct products of enzymatic reactions and provide a functional readout
of cellular state
\cite{fiehnMetabolomicsLinkGenotypes2002, pattiMetabolomicsApogeeOmics2012}.
Compared to genes and proteins that are regulated and
post-translationally modified, respectively, metabolites are most
predictive of the phenotype
\cite{raamsdonkFunctionalGenomicsStrategy2001}. So far, metabolomics
studies identified biomarkers for several diseases, and advanced our understanding of nutrition, the environment, and other complex biological systems.

%including
%pre-diabetes \cite{wang-sattlerNovelBiomarkersPrediabetes2012},
%diabetes \cite{suhreMetabolicFootprintDiabetes2010}, cancer
%\cite{spratlinClinicalApplicationsMetabolomics2009}, Parkinson's
%disease \cite{bogdanovMetabolomicProfilingDevelop2008}, Crohn's disease
%\cite{janssonMetabolomicsRevealsMetabolic2009}, and many others.

Mass Spectrometry (MS) techniques coupled with liquid or gas
chromatography separation techniques, LC-MS or GC-MS, have become a
standard analytical platform for untargeted metabolomics
\cite{alonsoAnalyticalMethodsUntargeted2015, kuehnbaumNewAdvancesSeparation2013}.
The LC or GC step aims to separate compounds within the sample, while
the MS step aims to ionize, fragment and detect a fragmentation pattern. For each compound, this pattern forms a \textit{spectral signature}, comprising a chromatographic retention time (RT) paired with mass
measurements (m/z) and their respective relative intensities for a particular compound and its fragments.

%there are now techniques for data processing (e.g., peak picking,
%missing value imputation, and adduct and degenerative feature removal)
%using tools such as XCMS \cite{mahieuRoadmapXCMSFamily2016},
%MarkerView™ Software, CAMERA \cite{kuhlCAMERAIntegratedStrategy2012},
%and creDBle \cite{mahieuSystemsLevelAnnotationMetabolomics2017}. These
%tools convert raw MS data into features. Each feature corresponds to a
%chemical compound, and is characterized using a spectral signature,
%comprising a chromatographic retention time (RT) paired with mass
%measurements (m/z) and relative intensities for a particular compound
%and its fragments.

Interpreting metabolomics measurements requires  \textit{annotation}, the process of assigning  putative
chemical identities to each spectral signature.  
From a
computational perspective, mapping molecules to their respective
spectral signatures represents a ``forward problem''. Mapping
spectral signatures back to their respective
molecular identity is an ``inverse problem''. The inverse problem is
exceptionally challenging as not all fragments are measured and many
isomers (same molecular formula but different atom configurations) have
almost indistinguishable  spectra.

Current annotation techniques attempt to solve the forward problem, and
can be broadly classified into two categories. Database lookup relies on comparing measured
spectra against experimentally generated spectra cataloged
in spectral databases such as METLIN
\cite{smithMETLINMetaboliteMass2005}, HMDB
\cite{wishartHMDBHumanMetabolome2007}, MassBank
\cite{MassBankPublicRepository}, and NIST
\cite{lamDevelopmentValidationSpectral2007}. Coverage of such libraries
however is limited, as experiments are required to generate signatures
from known, trusted chemical standards. Based on a predetermined set of \textit{candidate  molecules}, \textit{in-silico} annotation tools recommend a candidate molecule 
that best explains the measured spectra. The candidate set is typically culled from large molecular databases, such as PubChem, based on molecular mass or formula, if possible.  Earlier works generated
fragmentation patterns of candidate metabolites using rule-based
approaches, e.g., Mass Frontier \cite{bueschlNovelStableIsotope2014},
ACD/MS Fragmenter \cite{AccuratelyPredictMass}, and Hammer
\cite{zhouHAMMERAutomatedOperation2014}. Subsequent efforts introduced
combinatorial enumeration methods, e.g., MetFrag
\cite{wolfSilicoFragmentationComputer2010}, Fragment Identificator
(FiD) \cite{heinonenFiDSoftwareInitio2008}, Fragment Formula Calculator
\cite{wegnerFragmentFormulaCalculator2014}, and Mass Spectrum
Interpreter \cite{segataMetagenomicBiomarkerDiscovery2011}. More
recently, machine-learning algorithms have been investigated. CFM-ID
trains a probabilistic generative model of the 
fragmentation process to predict patterns of fragmentation
\cite{allenCFMIDWebServer2014, allenCompetitiveFragmentationModeling2015}.
CSI:FingerID \cite{duhrkopSearchingMolecularStructure2015} first
predicts a fragmentation tree based on a spectral signature
\cite{rascheComputingFragmentationTrees2011}. CSI:FingerID then uses
multiple-kernel learning
\cite{shenMetaboliteIdentificationMultiple2014} and support vector
machines (SVMs) to predict fragmentation tree properties that are then
searched against properties of candidate molecules. All of these techniques \textit{explicitly} model compound
fragmentation. A recent study proposed a model, NEIMS (Neural
Electron-Ionization Mass Spectrometry), to augment existing NIST 17 GC-MS libraries
with with synthetic spectra predicted from candidate molecules
\cite{weiRapidPredictionElectron2019}. The molecules are first mapped
to their ECFP (Extended-Connectivity Fingerprints) fingerprint that
record the count of molecular subgraphs within a specified radius
centered on each atom in the molecule. Using a radius of 2, the
fingerprint consisted of 4096 entries. The fingerprint is input to a 
fully connected feed forward neural network (FFNN) with a gated
bidirectional design to improve the prediction accuracy. However, fingerprints,
while common and useful, are not tailored to the prediction task. Moreover, NEIMS was only evaluated on the NIST data generated via the GC-MS technique.   GC-MS fragmentation patterns are simpler than those obtained using LC-MS. Further GC-MS is  typically used to measure  volatile compounds (or compounds  that can be extracted into an organic solvent and vaporized using GC), which typically have masses less than 500 Daltons.

Here, we evaluate the feasibility of using Graph Neural Networks
(GNNs) on the MS spectral prediction task. GNNs have been shown powerful
in terms of learning representations from structured data
\cite{gilmerNeuralMessagePassing2017, kipfSemiSupervisedClassificationGraph2017, velickovicGraphAttentionNetworks2018, wuComprehensiveSurveyGraph2020},
such as social networks, knowledge graphs and  molecules. Here,
we represent each molecule as a graph, where atoms are represented as nodes, and bonds are represented as edges. We explore the use of both Graph Convolutional Networks
(GCN) \cite{kipfSemiSupervisedClassificationGraph2017} and Graph
Attention Networks (GAT) \cite{velickovicGraphAttentionNetworks2018}. We train and evaluate our technique on the NIST 17 LC-MS dataset.

\hypertarget{methods}{%
\section{Methods}\label{methods}}

\begin{wrapfigure}{R}{.3\textwidth}
\begin{center}

\includegraphics{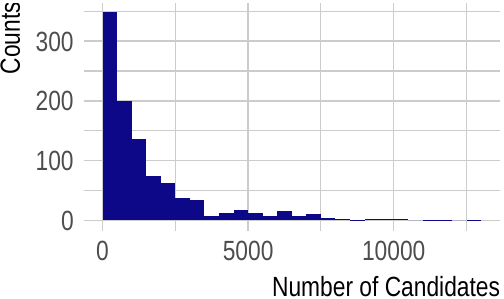}

\caption{Distribution of number of candidates for test molecules. The average number of
candidates per test molecule is 1,530.}
\label{fig:distribution}
\end{center}
\end{wrapfigure}

\hypertarget{datasets}{%
\subsection{Datasets}\label{datasets}}

The NIST 17 dataset consists of several mass spectrum datasets, including the EI MS library, the GC library, and the tandem MS/MS library. For training and evaluation, we focused our efforts on the tandem MS/MS dataset, and selected spectra obtained using HCD  (higher energy collisional dissociation), which 
provides a richer and more varied spectra than CID (collision induced dissociation). The HCD option was used to measure 69.6\% of the MS/MS data.
We also restricted the precursor types to only include ones that are common in the dataset (e.g., {[}M+H{]}+, {[}M-H{]}-, {[}M+H-H2O{]}+, {[}M-H-H2O{]}, etc.). We selected one spectrum with the largest collision energy under 40eV for each training and test molecule.

From the reduced dataset, we  randomly select 1000 molecules as a test set. We then split the remaining dataset into training and
validation (4:1 ratio), yielding  6,188 training and 1,539 test molecules. The validation set is used for model selection while the test set is used to report performance. The relevant candidate sets for the test molecules were queried from 
PubChem  using the exact molecular formula of each test molecule
 (Figure
\ref{fig:distribution}).

\hypertarget{data-preprocessing}{%
\subsection{Data Preprocessing}\label{data-preprocessing}}
A molecule is represented as a graph \(G = (V, E)\), where  atoms correspond to  the node set \(V\) and bonds correspond to the edge set \(E\). Let
\(\bX = \{x_1,x_2,...,x_N\}, x_i \in \bbR^F\) denote a set of node
features, where \(N\) is the number of nodes and \(F\) is the number of
node features. The connectivity among nodes is described by an adjacency
% Hao - what is \bbN? 
matrix \(\bA \in \{0, 1\}^{N \times N}\). Node features
include  standard atomic weight, atom type, number of bonds, 
number of neighboring hydrogen atoms, and  indicators if the atom is part of a ring or an aromatic ring. 

%We also tested if including edge type as node attribute will contribute to the performance of this model.

The spectra data is a list of paired mass-to-charge ratio (m/z) and their  relative intensities. Each m/z value was rounded down to the nearest integer m/z bin. If more than one m/z values are rounded to the same bin, we record the highest intensity. The range of intensity values has a long tail of large values, so we take either the logarithm or the square-root of these intensity values as the fitting target. Then denote the vector as \(\by\).

%A spectrum is an array of relative intensities \(\by\) at each integer
%mass-to-charge ratio (m/z). As MS/MS is typically utilized for measuring molecules with masses less than 1000 Dalton, and the charge z is typically +/- 1, we only included molecules whose largest spectra was less than 1000. 
%Therefore, \(\by \in \bbR^{1000}\). As the intensities of the spectra peaks were normalized relative to the peak with the highest intensity, which was set to 999.0, the spectra intensities ranged from 0 to 999.0.
%We explored using the log transformation and the square-root transformation to scale the peak intensity.
%Previous research suggested that  data \cite{steinOptimizationTestingMass1994}. 

\hypertarget{neural-network}{%
\subsection{Neural Network}\label{neural-network}}

\begin{figure}[t]
\centering
\includegraphics[scale=0.8]{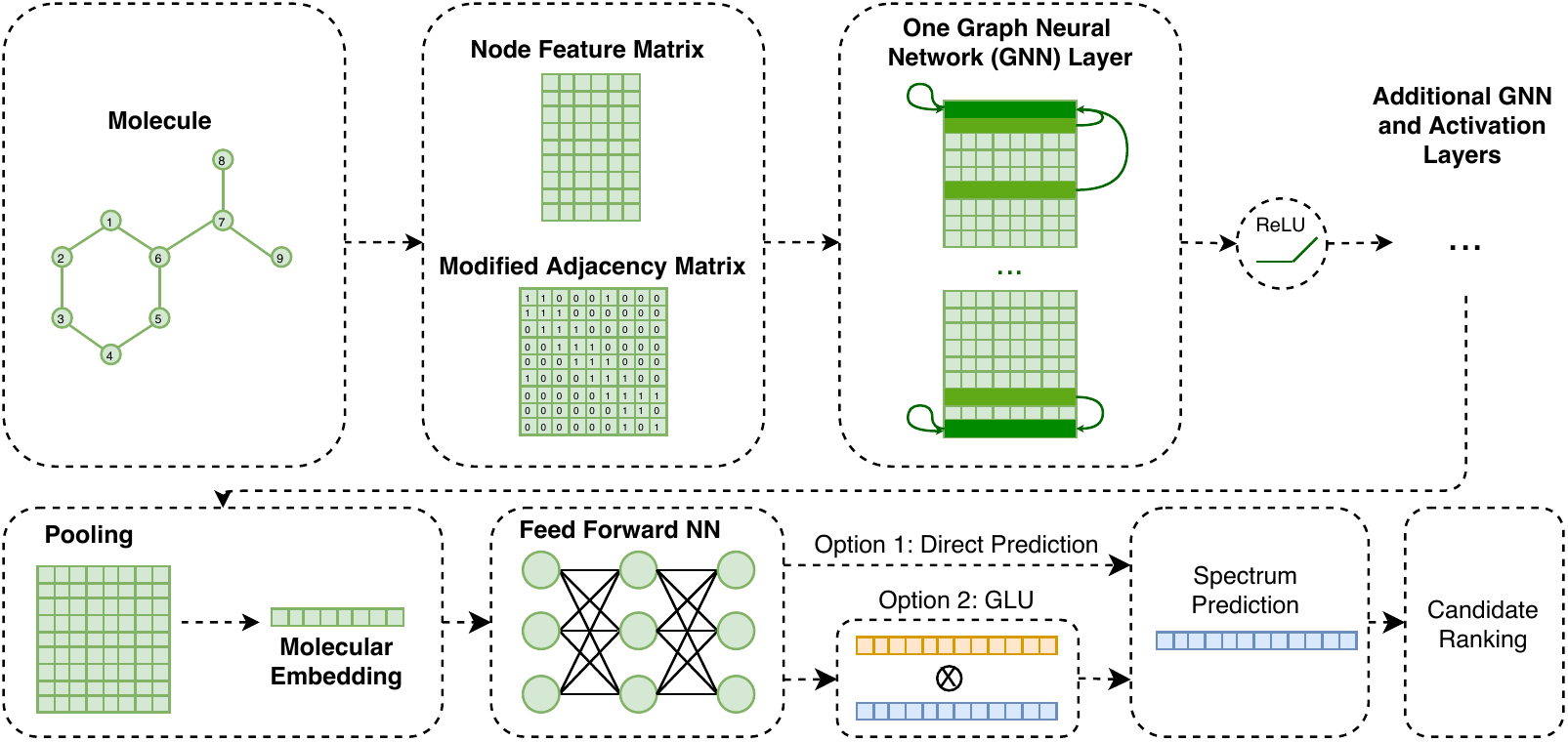}
\caption{Illustration of the network architecture used in this study}
\label{fig:flowchart}
\end{figure}

An overview of our neural network (NN) architecture is illustrated in Figure
\ref{fig:flowchart}. The model comprises multiple GNN layers, a pooling layer and 
a fully-connected feed forward regression model. In each GNN layer, node information is propagated
along graph edges. More specifically, let \(\bh = \{h_1, h_2,...h_N\}, h_i \in \bbR^H\) denote 
the input of a GNN layer, where \(H\) is the dimension of input node embedding vector. Let \(\bh'\) 
denote the output with \(H'\) as the dimension of the new node embedding. The weight matrix of such a 
layer is \(\bW \in \bbR^{H' \times H}\) and the bias term is \(b\). In the first
GNN layer, \(\bh\) is  set to the original node features, \(\bX\).
By stacking several layers of GNN layers, information on each node
propagates along edges to a broader neighborhood within the graph.

We explore using two  GNN implementations: GCN
\cite{kipfSemiSupervisedClassificationGraph2017} and GAT
\cite{velickovicGraphAttentionNetworks2018}. The  propagation rule of a single GCN layer is given as:

\setlength{\belowdisplayskip}{0pt}
\setlength{\belowdisplayshortskip}{0pt}
\setlength{\abovedisplayskip}{0pt}
\setlength{\abovedisplayshortskip}{0pt}

\[
h_i' = \sigma(\sum_{j \in \calN(i)} \frac{1}{c_{ij}} \bW h_j + b)
\]

where \(\calN(i)\) is the set of neighbors of node \(i\) and \(\sigma\)
is an activation function. \(c_{ij}\) is a normalization term and
it is set equal to \(\sqrt{|\calN(i)||\calN(j)|}\). The normalization
penalizes nodes with too many connections to avoid
extreme values. In contrast to GCN, GAT utilizes attention mechanism  to propagate information as follows:
\[
h_i' = \sum_{j \in \calN(i)} \alpha_{ij} \bW h_j
\]

where \(\alpha_{ij}\) is the attention term and equals to
\(\mathrm{softmax}_i(\sigma(\vec{a}^T[\bW h_i||\bW h_j]))\). In essence, GAT introduces  additional
trainable attention weights, $\vec{a}^T$, on the concatenated \(Wh_i\) and \(Wh_j\) vectors, to model a
 weighting term that controls how the message 
\(Wh_j\) from each neighbor \(j\) should be propagated to node \(i\). The softmax activation
ensures that the sum of these weighting terms equals to 1 for each node. By design, GAT does not have a bias term.

In the second step of this model, after the learned node representation \(\bh\)
is obtained,  node information is ``pooled'' into a graph embedding
vector \(\bv \in \bbR^{H}\), thus aggregating information about the entire graph.
We compared several different pooling methods including Global
Maximum, Global Average and Global Attention
\cite{liGatedGraphSequence2017}.

After the pooling layer, the graph embedding vector
\(\bv\) is fed into a feed forward network (FFNN) that predicts
 \(\hat{\by} \in \bbR^{1000}\). We also evaluated a gated linear unit (GLU) \cite{dauphinLanguageModelingGated2017} instead of a dense layer for the outcome prediction. A GLU predicts two outputs simultaneously while one of the two outputs is activated by sigmoid and acts as a gate for the other output. The final output is activated by a ReLU function \cite{relu} to ensure positive outcome. 

\hypertarget{training}{%
\subsection{Training}\label{training}}

The model was trained  by minimizing the mean square error (MSE) between \(\hat{\by}\) and \(\by\). Since spectrum intensity itself is 
relative, we normalized
both \(\by\) and \(\hat{\by}\). 
%Our best performing model was trained using a learning rate at 5E-8. 
To reduce overfitting, we used L2 regularization
with lambda set to 1.0 and a dropout rate at 0.5. All models
were trained on a Nvidia P100  for a maximum of 1,000 epochs
using Adam \cite{kingmaAdamMethodStochastic2017} with early stopping on
validation loss with a window size of 15.

\hypertarget{evaluation-and-study-baselines}{%
\subsection{Evaluation}\label{evaluation-and-study-baselines}}

We evaluated our models using two metrics. First, the cosine similarity
between the predicted and target spectra is used to assess the quality of the predicted spectra.
Second, recall@k, a common metric for evaluating annotation tools, 
measures the the portion of correctly identified molecular identities in the test dataset when considering 
the top k ranked candidate molecules for each test spectra. We used the NEIMS model \cite{weiRapidPredictionElectron2019} as a baseline.  

%We also applied
%our best-performing model on the GNPS dataset, which was used in the
%CSI:FingerID/SIRIUS paper \cite{duhrkopSIRIUSRapidTool2019}.

\pagebreak

\hypertarget{results}{%
\section{Results}\label{results}}

\setlength{\tabcolsep}{3pt}
\begin{wraptable}{r}{4in}
\caption{Summary of sesults in cosine similarity and recall@k}
\fontsize{9}{10}\selectfont
\begin{tabular}{lccccc}
\toprule
Experiment & Similarity & Avg Rank & Recall@1 & Recall@5 & Recall@10\\
\midrule
\multicolumn{6}{l}{NIST 17 + Candidates from PubChem (Sampled with Average Size = 50)} \\
\hspace{1em}NEIMS & 0.157 & 16.7 & 30.2 & 52.5 & 65.1 \\
\hspace{1em}GCN (3 layers) & 0.426 & 10.8 & 33.9 & 62.3 & 75.9 \\
\hspace{1em}GAT (3 layers) & 0.468 & 9.2 & 36.0 & 65.2 & 76.6 \\
\hspace{1em}GAT (10 layers) & 0.512 & 7.1 & 41.2 & 70.7 & 81.4 \\
\hspace{1em}GAT + GLU & \textbf{0.517} & \textbf{6.8} & \textbf{45.1} & \textbf{70.8} & \textbf{83.3} \\
%GAT + GLU on GNPS & 0.242 & 26.0 & 19.4 & 36.8 & 50.1 \\

\bottomrule
\end{tabular}
\end{wraptable}
Our experiments had two purposes: 1) explore a wide variety of NN architectures and select the  model that minimizes the MSE, and 2) evaluate the model's ranking performance on the candidate set as compared to our baseline model.

Compared with the molecular fingerprint based NEIMS model (Table 1), GNN-based NN models generate
significantly more accurate spectra and exhibit  higher recall@k. To
find  the  NN architecture that minimizes the MSE, we examined the effect of including adjacent bond information in the model. We also performed hyper-parameter
searching on MS intensity data transformation methods (square
root or log transformation), graph pooling methods on nodes (Global Max,
Global Average or Global Attention), GNN types (GCN or GAT), 
GNN architecture and the outcome prediction layer (Dense layer or GLU).
Including bond information was beneficial. Log transformation  consistently provides better performance than square-root transformation. GAT performs better
than GCN, suggesting that the attention mechanism is 
effective for this task. We also 
observed that the performance of GCN quickly dropped as the number of
GCN layers increase. This agrees with previous reported findings
\cite{wuComprehensiveSurveyGraph2020}. However, by
using GAT, it was possible to build deeper GNN
models. The best performance happens when there are 10 layers of stacked GATs
with 64 hidden units, with the output predicted by GLU. Among the three pooling methods, Global Max provided
better performance than the other two. After the model was trained, it
achieved \textasciitilde11,000 predictions per second on an Nvidia
V100 GPU card.

\begin{wrapfigure}[25]{l}{4in}

\includegraphics{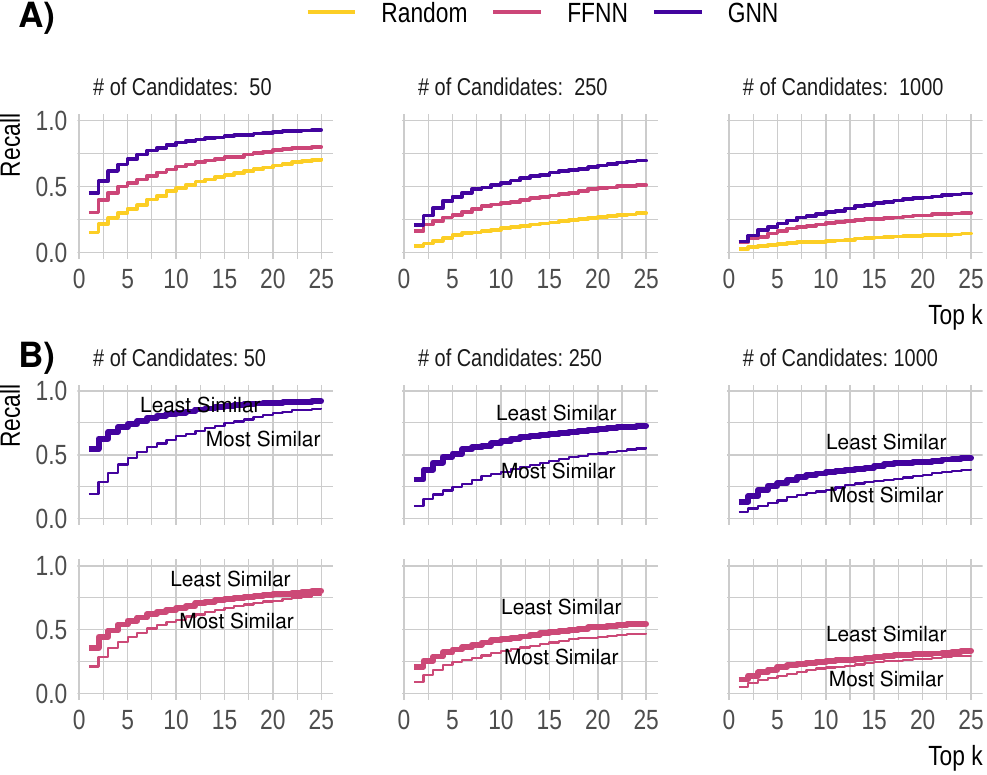}

\caption{Impact of: (A) candidate size and (B) candidate similarity on ranking results}
\label{fig:candidatesize}
\end{wrapfigure}
To evaluate how the size of candidate set impacts ranking performance, we sample a proportion of candidates for each test molecule. We calculated the ratio of the average number of candidates per test molecule (1,530) and 50, 250 and 1000 and used these ratios to scale the candidate set for each test molecule appropriately. As shown in
Figure 3A, the recall@k performances for both models decrease as 
the average number of candidates increases. With more
candidates to choose from, correct identification becomes more challenging. In Figure 3B, where MACCS Fingerprints were used to identify the
most similar and least similar compounds in the candidate set, recall@k performances are low when candidate molecules are similar with their target molecule. 

Ranking on a candidate set using recall@k is widely used for annotation evaluation. Current evaluation datasets (in terms of test molecules and their candidate sets), however, vary tremendously. Our results show that ranking results for both models heavily depend on the candidate set size and the similarity of the candidates to the target molecule, thus highlighting the need for better and consistent evaluation protocols for annotation tools.

\hypertarget{conclusion}{%
\section{Conclusion}\label{conclusion}}

We investigated several GNN-based models to predict the mass spectra for query  molecular structure. Our model outperforms previously reported NN
models. Importantly, we  found that
ranking results are heavily dependent on the candidate set size as well as the similarity of candidate molecules with target
molecule. We encourage researchers to standardize  performance evaluation for the MS spectra prediction task, and to consider GNN-based methods for annotation.

\pagebreak

\bibliography{references.bib}

\end{document}